%% file: main.tex
\documentclass{article}
\usepackage{iclr2027_conference,times}
\usepackage{amsmath,amssymb,amsfonts,amsthm,booktabs,multirow,array,graphicx,xcolor,colortbl}
\usepackage[utf8]{inputenc}
\usepackage[T1]{fontenc}
\usepackage{url,microtype,xspace,caption,enumitem,placeins}
\input{macros}
\title{OC-GS: Gaussian Splatting for\\
Irregular Turntable Capture}
\author{Jae Joong Lee \& Bedrich Benes \\
Department of Computer Science\\
Purdue University\\
\texttt{\{lee2161,bbenes\}@purdue.edu}}
\iclrfinalcopy
\begin{document}
\maketitle
\lhead{}
\renewcommand{\topfraction}{0.9}
\renewcommand{\bottomfraction}{0.8}
\renewcommand{\textfraction}{0.08}
\renewcommand{\floatpagefraction}{0.8}
\input{src/0_abstract}
\input{src/1_intro}
\input{src/3_method}
\input{src/4_experiments}

\FloatBarrier
\input{src/5_limitations}
\input{src/6_conclusion}
\input{src/2_related}
\bibliography{bibliography}
\bibliographystyle{iclr2027_conference}
\clearpage
\appendix
\setcounter{table}{0}
\setcounter{figure}{0}
\setcounter{equation}{0}
\renewcommand{\thetable}{A\arabic{table}}
\renewcommand{\thefigure}{A\arabic{figure}}
\renewcommand{\theequation}{A\arabic{equation}}
\input{src/7_appendix}
\end{document}

%% file: macros.tex
\newcommand{\name}{OC-GS\xspace}

\newcommand{\figref}[1]{Fig.~\ref{#1}}
\newcommand{\tabref}[1]{Tab.~\ref{#1}}
\newcommand{\secref}[1]{Sec.~\ref{#1}}
\newcommand{\appref}[1]{App.~\ref{#1}}
\newcommand{\apptabref}[1]{Tab.~\ref{#1}}

\definecolor{OurRow}{HTML}{ECECEC}
\newcommand{\ourrow}{\rowcolor{OurRow}}
\newcommand{\best}[1]{\textbf{#1}}
\newcommand{\second}[1]{\underline{#1}}

%% file: src/0_abstract.tex
\begin{abstract}
Uneven rotation and dropped frames make equal-angle assumptions unreliable for turntable reconstruction. We present OC-GS, an object-centric Gaussian splatting that refines each image's angle while maintaining a shared camera, rotation axis, and pivot. This \emph{orbit-consistent refinement} jointly optimizes image-derived geometry and angles to reconstruct objects from sparse, irregular captures. On rendered objects with 12, 8, and 6 irregularly spaced views, OC-GS achieves mean foreground PSNR scores of 21.26, 19.36, and 15.83dB, respectively, exceeding all four evaluated pose-free Gaussian splatting baselines in each condition. Under a shared trainer, refining image-estimated angles improves mean foreground PSNR by 7.88dB over keeping those estimates fixed. An ablation study shows that both image-derived angle initialization and the shared motion model contribute to the improvement. On real captures, OC-GS's refinement increases mean foreground PSNR by 0.70dB. Results show that refining uncertain angles within a shared motion model improves reconstruction from sparse, irregular turntable captures.
\end{abstract}

%% file: src/1_intro.tex
\section{Introduction}
\label{sec:intro}
A turntable scanning setup is a common method that lets a fixed camera capture an object from multiple directions. It is used in plant phenotyping~\citep{yang2024plant}, cultural heritage digitization~\citep{marshall2019artefacts}, biological specimen digitization~\citep{stroebel2018disc3d}, and industrial surface inspection~\citep{kaichi2018inspection}. Scanning time is critical, as some systems scan hundreds of objects a day. Reconstructing objects from fewer images can reduce capture time, but it complicates the reconstruction. Uneven rotation and dropped frames make inter-image angles unequal. We study this setting, where image order is known, but rotation angles are not.

Recent methods reconstruct Gaussian scenes without supplied camera poses. AnySplat~\citep{jiang2025anysplat}, InstantSplat~\citep{fan2024instantsplat}, and SPFSplat~\citep{huang2025spfsplat} use learned geometry to reconstruct from images. RotGS~\citep{kim2026rotgs} incorporates a shared rotation axis and optical-flow supervision for turntable capture. These methods provide useful geometry and motion estimates, but irregular angular spacing remains a challenge.

A common assumption assigns angle $360^\circ j/N$ to view $j$, treating frame index as a measure of rotation. RotGS uses this schedule with an interpolated endpoint correction~\citep{kim2026rotgs}, which cannot represent arbitrary gaps between frames. Image-derived angles can capture unequal spacing, but keeping inaccurate estimates fixed preserves their errors. Independently optimizing every camera allows corrections that may violate the turntable motion. We instead refine individual angles within one shared rotation model.

We propose \name, object-centric Gaussian splatting with \emph{orbit-consistent refinement} (\figref{fig:overview}). Image predictions initialize the geometry and angles from a pre-trained initializer model such as VGGT~\citep{wang2025vggt}. OC-GS jointly optimizes the Gaussian scene, one camera, one rotation axis, one pivot, and an angle correction for each view. The shared rig maintains a consistent rotation trajectory, while the corrections adjust unequal angular gaps. Reconstruction therefore uses the captured images without requiring measured angles or constant rotation speed.

\input{figures/tex/overview}

Our contributions are:
\begin{itemize}[leftmargin=1.5em,itemsep=1pt,topsep=2pt]
\item \textbf{Image-derived initialization for irregular capture:}
predicted geometry to initialize the scene and fit a circle to predicted camera centers to estimate non-uniform angles (\secref{sec:angle_sources}).
\item \textbf{Per-view corrections to angular spacing:}
we learn a correction for each view and subtract the mean correction to adjust relative spacing without introducing a redundant common rotation (\secref{sec:angle_sources}).
\item \textbf{Orbit-consistent joint refinement of shape and motion:}
we optimize the angles, Gaussian scene, camera, axis, and pivot together under the turntable constraint. Component ablations show improvements over fixed angle estimates (\secref{sec:synthetic_benchmark}).
\end{itemize}

%% file: figures/tex/overview.tex
\begin{figure}[t]
\centering
\includegraphics[width=\textwidth]{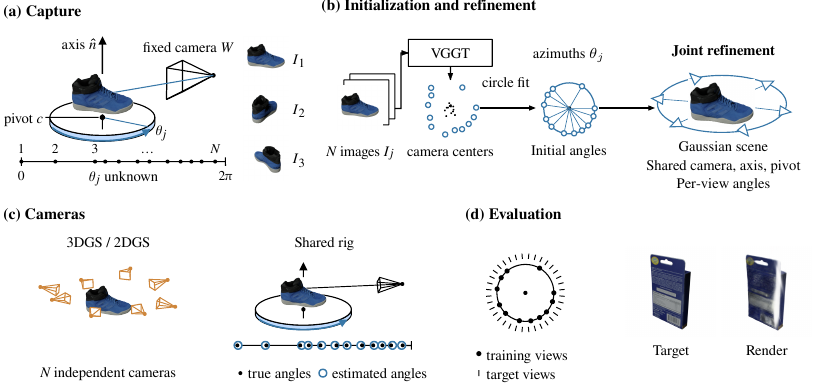}
\caption{\textbf{OC-GS overview.} (a)~A fixed camera captures an object rotating about one axis with unequal angular gaps. (b)~VGGT predicts initial geometry and cameras. A circle fit to the camera centers gives starting angles. OC-GS jointly refines the Gaussian scene, shared camera, axis, pivot, and per-view angles. (c)~The shared turntable rig (blue) constrains the effective cameras, unlike independent cameras (orange). (d)~Reconstruction is evaluated by comparing rendered and target images.}
\label{fig:overview}
\end{figure}

%% file: src/3_method.tex
\section{OC-GS: orbit-consistent refinement}
\label{sec:method}\label{sec:intervention}
OC-GS refines image-derived geometry and angles under a shared turntable model. Each image has its own rotation angle, while all views share one camera, rotation axis, and pivot.

\subsection{Turntable rig}
\label{sec:capture}\label{sec:ocgs}
We take $N$ ordered RGB images $I_j$, foreground masks, and camera intrinsics $K$ of an object completing a turn about a fixed axis. The angle of each image is unknown. Our goal is to recover a canonical Gaussian object $\mathcal G$ and cameras for novel-view rendering. We do not evaluate segmentation or intrinsic calibration.

Let $W$ be the physical camera's world-to-camera transform, $\hat n$ the unit rotation-axis direction, and $c$ a pivot on that axis, expressed in world coordinates. For image~$I_j$, the object rotates by~$\theta_j$. Using $4\times4$ homogeneous transforms acting on column vectors, the object motion~$A_j$ and effective object-to-camera transform~$W_j$ are
\begin{equation}
 A_j=T(c)R(\hat n,\theta_j)T(-c),\qquad W_j=W A_j.
 \label{eq:rig}
\end{equation}
where $T(v)$ denotes translation by $v$, and $R(\hat n,\theta_j)$ denotes rotation about the axis through the origin in direction $\hat n$. The product first moves the pivot to the origin, then rotates the object and translates it back. $W$ maps the result into camera coordinates. Gaussian centers and local frames share the same rigid motion (\appref{app:protocol}). This model groups all effective cameras into one trajectory while allowing unequal angular gaps.

\subsection{Angle sources and spacing errors}
\label{sec:angle_sources}
\paragraph{Initial angles.}
For experiments on synthetic objects, we use VGGT~\citep{wang2025vggt} to initialize both geometry and cameras, keeping its predictions fixed across variants to isolate the effect of angle refinement. We estimate angles by fitting a circle to the predicted camera centers, using the first image to define the reference angle (\appref{app:protocol}). For real captures, we use AnySplat-derived angles~\citep{jiang2025anysplat}. OC-GS can use any initializers, but we use the best-performing ones (Sec.~\ref{sec:initializers}).

\paragraph{Angle refinement.}
OC-GS learns an angle correction for each image to improve the estimated spacing. Adding the same correction to every angle would only rotate the object, so we subtract the mean correction to adjust relative angles. We jointly optimize the corrections with the Gaussian scene, camera, axis, and pivot, while keeping the intrinsics fixed. Training uses an~$L_1$ photometric loss with opacity and scale regularization. Initialization and optimization use only the input images (\appref{app:protocol}).

\subsection{Comparison variants.}
We compare five settings in the same 2DGS~\citep{huang20242d} to distinguish the effects of angle refinement and shared motion. The \textbf{independent-camera variant} optimizes each camera separately. Within the shared rig, the \textbf{uniform-angle variant} assumes equal spacing, the \textbf{frozen-angle variant} keeps image-derived angles fixed, and \textbf{OC-GS} refines them. The \textbf{true-angle reference} uses exact angles but still estimates the axis and pivot.

We also use various spacings and image counts to validate the method. \textbf{Regular-12} and \textbf{Irregular-12} each contain twelve views, with equal or unequal angular gaps, respectively. \textbf{Irregular-8} and \textbf{Irregular-6} use fewer views with unequal gaps. Each configuration runs for 30k steps with three random seeds (\appref{app:protocol}).

\paragraph{Evaluation} evaluates reconstruction quality on 36 target views. We measure foreground pixel accuracy with FG-PSNR, perceptual similarity with LPIPS~\citep{zhang2018lpips}, and silhouette overlap with IoU. Relative rotation error (RRE) compares predicted and reference rotations between camera pairs. For each object and capture condition, we average within runs and then across three different seeds. Metric definitions are given in \appref{app:alignment}.

%% file: src/4_experiments.tex
\section{Results}
\label{sec:experiments}\label{sec:historical_factorial}
We evaluate reconstruction quality, the contribution of angle refinement, and computational cost. We compare OC-GS with existing methods, then examine its components in ablation experiments on synthetic objects. We also test refinement on real captures and introduce known angle errors to determine when it helps. Finally, we study the reasons for limited baseline performance and compare runtime and memory use.
\input{figures/tex/claim_support}

\subsection{Comparison with existing reconstruction methods}
\label{sec:external_baselines}
We compare \name with five baseline methods and two calibrated references on 100 objects from Google Scanned Objects~\citep{downs2022google}. The four capture conditions use 12 regularly spaced views or 12, 8, or 6 irregularly spaced views. Within each condition, all methods are evaluated on the same 36 target views (\appref{app:corrections}).

The baselines are AnySplat~\citep{jiang2025anysplat}, SPFSplat~\citep{huang2025spfsplat}, InstantSplat~\citep{fan2024instantsplat}, RotGS~\citep{kim2026rotgs}, and Nerfacto~\citep{tancik2023nerfstudio}. We use the released versions of AnySplat, SPFSplat, InstantSplat, and RotGS. Nerfacto also runs for 30k steps, starting from cameras predicted by VGGT~\citep{wang2025vggt} for the input views. Moreover, to evaluate two neural scene representation models such as Nerfacto and Splatfacto, we provide the true training cameras to run their published pipelines.

\paragraph{OC-GS leads under irregular capture.}
With 12, 8, and 6 irregularly spaced views, OC-GS achieved mean foreground PSNR scores of 21.26, 19.36, and 15.83dB, respectively, outperforming all evaluated pose-free GS methods at each view count (\figref{fig:claim_support}). When both methods yield valid results, OC-GS exceeded RotGS by at least 10.25dB across these conditions and AnySplat by 4.11dB with six views (\apptabref{tab:external_paired}). Our method, OC-GS, achieves the highest FG-PSNR across all irregular image settings compared with other baselines, and it scores, on average, 54.5\% better than AnySplat, which has the second-best performance.
Reconstruction quality decreased as fewer views were available, but OC-GS remained ahead of these baselines. 

The qualitative examples show similar performance gaps (\figref{fig:external_qualitative}). OC-GS preserves the bull's leg positions and silhouette, while several baselines produce distorted or displaced reconstructions. For the Bathroom Set, OC-GS reconstructs the main parts of the scene, while baseline results show missing parts or scattered visuals.

\input{figures/tex/external_qualitative}

\paragraph{Comparison with calibrated references.}
Comparisons with the two neural scene representations show the same trend.
OC-GS outperforms Splatfacto with true cameras at eight and six irregular views. With twelve views, Splatfacto performs better under both regular and irregular capture. Across cases with valid camera alignment for both methods, OC-GS achieves 1.43dB higher mean FG-PSNR and 0.103 lower mean object-crop LPIPS (\apptabref{tab:external_pooled}).

\input{src/4_revision_results}

\subsection{Real-capture evidence and initialization}
\label{sec:real_evidence}
\label{sec:real}
We next validate whether angle refinement also improves reconstruction on the RotGS released dataset, which consists of real captures. We compare the initial angle estimates and optimized angles during the reconstruction using 24, 17, 13, and 7 training views.
Both settings use the same initial angles estimated by AnySplat. We show that optimizing these angles improves mean FG-PSNR by 0.70dB and lowers mean LPIPS at every view count (\appref{app:real_profile}). 

\paragraph{Initializers on synthetic and real captures.}
\label{sec:initializers}
We further study the effectiveness of different initializers with OC-GS on synthetic and real captured datasets. VGGT shows the best performance, which outperforms the second-best, VGGT-$\Omega$, by 0.86 dB (\tabref{tab:ocgs_initializers}). On the RotGS released dataset, AnySplat outperforms the second-best $\pi^3$ by 0.36dB (\tabref{tab:init_sources}). 

\input{tables/tex/ocgs_initializers}
\input{tables/tex/init_sources}

\subsection{When does angular refinement help?}
\label{sec:dose}
Because we use pretrained models for the angle, they can be inaccurate due to the input image or the initialization method. To ablate those effects, we study the effect of angle error directly by adding noise of known magnitude to the ground-truth angles. We then compare keeping the noisy angles unchanged with optimizing them during reconstruction. Both settings use the same twelve regularly spaced input views, VGGT-initialized geometry, and an estimated rotation axis.

We remove the average angular offset introduced by the noise, leaving only errors in the relative angles between views. We measure these errors using their root-mean-square (RMS) value and test levels from $0^\circ$ (exact angles) to $10^\circ$ (\tabref{tab:followup_dose}).

\input{tables/tex/followup_dose}

Angle refinement improves reconstruction at all tested error levels from $0.5^\circ$ to $10^\circ$ RMS (\tabref{tab:followup_dose}). Compared with keeping the initial angles unchanged, the FG-PSNR gain improves from 4.75dB at $0.5^\circ$ to 11.25dB at $10^\circ$. At the largest error, refinement raises FG-PSNR from 9.08dB to 20.33dB. These results support refining imperfect angle estimates in our reconstruction setting. When the initial angles are exact, keeping them unchanged improves performance by 3.16dB (\appref{app:followup}).

Also, with exact initial angles, both variants fit the training images similarly well, but refinement produces worse results on unseen views. This is consistent with overfitting and shows why training-image quality alone is insufficient to assess reconstruction (\appref{app:followup}).

\subsection{Sources of error in the baseline reconstructions}
\label{sec:baseline_diagnostics}
We run additional experiments to investigate why the baselines perform worse on unseen views. These validate camera consistency, the ability to reproduce training images, and the spatial distribution of reconstructed geometry.

\paragraph{RotGS: sensitivity to the rotation axis.}
RotGS reproduces its training images well when rendered through its estimated cameras and is also designed for the turntable setup, but quality decreases when the same scene is rendered through aligned reference cameras. In a separate regular-capture experiment on five objects, fixing the rotation axis to its true value increases mean FG-PSNR from 15.97 to 29.45dB, with improvements on every object. This result suggests that axis estimation can limit reconstruction even when the input views are regularly spaced (\appref{app:rotgs_axis_audit}).

\paragraph{Image-based baselines: camera errors and limited training fit.}
For AnySplat and InstantSplat, replacing inferred camera parameters with calibrated ones while holding the reconstructed scene fixed reduces training-image quality. This indicates inconsistency between the reconstructed scene and the true cameras. SPFSplat also struggles to fit the training images, even with calibrated intrinsics. Correcting InstantSplat's camera export improves target-view quality only slightly (\appref{app:baseline_failure}).

\paragraph{Nerfacto: good training fit, limited unseen views quality.}
Nerfacto is able to fit sparse training images well while producing unseen views of limited visual quality, even with true cameras. In the reconstructions studied here, much of the predicted content lies outside the object region. In a four-object experiment with true cameras and known scene bounds, replacing scene contraction with a bounded field increases mean target FG-PSNR from 6.80 to 17.78dB under irregular capture.

\paragraph{Runtime and memory use.}
We also evaluate the cost comparison. OC-GS achieves the highest mean FG-PSNR and uses the least peak GPU memory, 1.51 GB (\tabref{tab:cost}). Fitting takes 16.2 minutes, less than RotGS and Nerfacto. InstantSplat and the feed-forward methods are faster but have lower reconstruction quality. VGGT inference is precomputed for OC-GS.

\input{tables/tex/cost}

%% file: figures/tex/claim_support.tex
\begin{figure}[t]
\centering
\includegraphics[width=\linewidth]{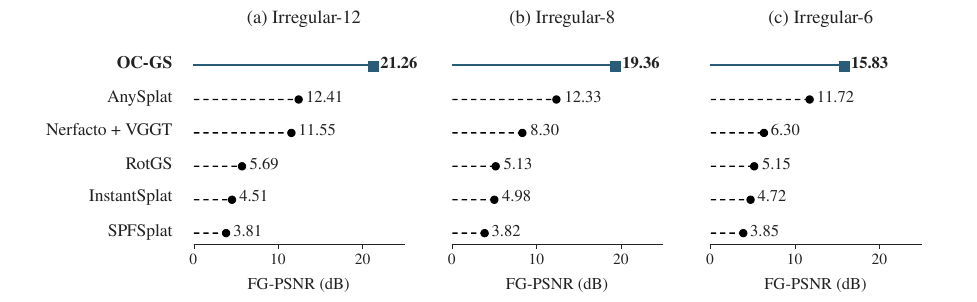}
\caption{\textbf{Reconstruction quality under irregular capture.} Mean target FG-PSNR with 12, 8, and 6 views (higher is better). Objects must have valid camera alignment in all five component configurations and all three seeds. Each baseline's alignment failures are also excluded. Methods retain their own training settings.}
\label{fig:claim_support}
\end{figure}

%% file: figures/tex/external_qualitative.tex
\begin{figure}[t]
\centering
\input{figures/external_qualitative_tex/grid}
\caption{\textbf{Selected reconstructions under irregular capture.} Hereford Bull with six views and Bathroom Set with twelve, shown at the first target view with the same crop across methods. SPFSplat has no valid placement for Hereford Bull.}
\label{fig:external_qualitative}
\end{figure}
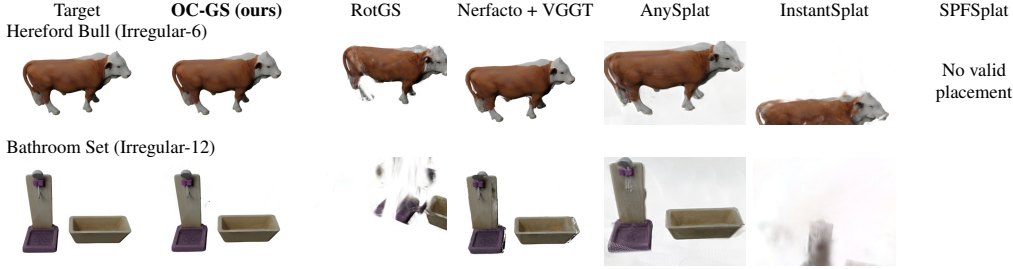

%% file: figures/external_qualitative_tex/grid.tex
\begingroup
\setlength{\tabcolsep}{1.5pt}
\renewcommand{\arraystretch}{1.0}
\scriptsize
\begin{tabular}{@{}ccccccc@{}}
Target & \textbf{OC-GS (ours)} & RotGS & Nerfacto + VGGT & AnySplat & InstantSplat & SPFSplat \\
\multicolumn{7}{@{}l}{\rule{0pt}{1.5ex}Hereford Bull (Irregular-6)} \\[-1pt]
\includegraphics[width=.134\linewidth]{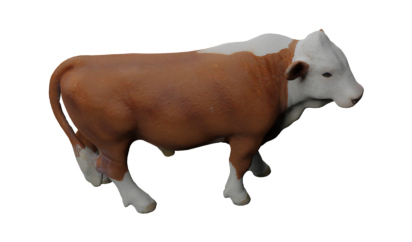} & \includegraphics[width=.134\linewidth]{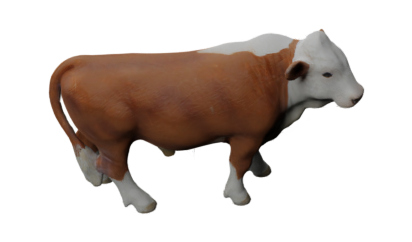} & \includegraphics[width=.134\linewidth]{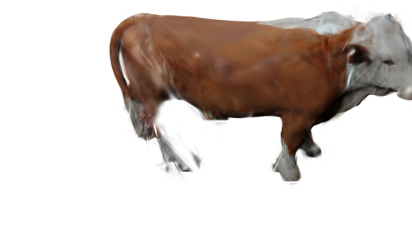} & \includegraphics[width=.134\linewidth]{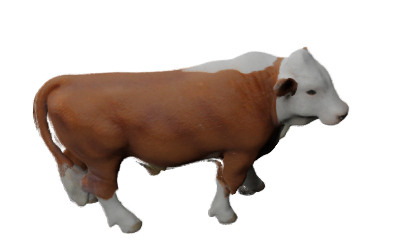} & \includegraphics[width=.134\linewidth]{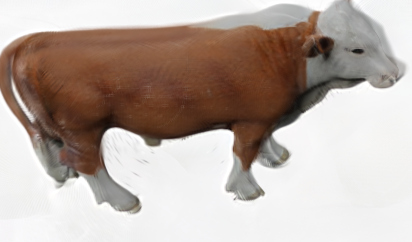} & \includegraphics[width=.134\linewidth]{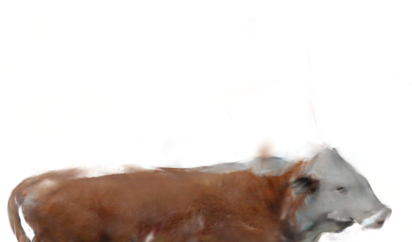} & \parbox[b][0.07871\linewidth][c]{.134\linewidth}{\centering No valid\\placement} \\[3pt]
\multicolumn{7}{@{}l}{\rule{0pt}{1.5ex}Bathroom Set (Irregular-12)} \\[-1pt]
\includegraphics[width=.134\linewidth]{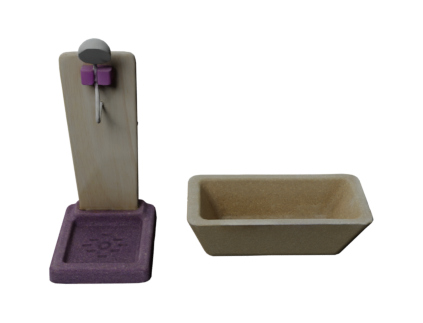} & \includegraphics[width=.134\linewidth]{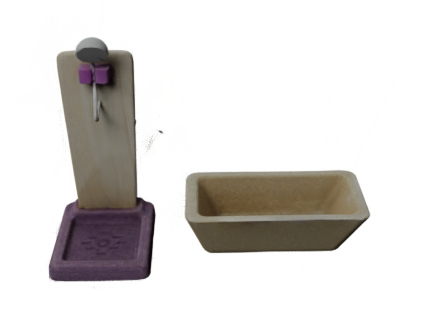} & \includegraphics[width=.134\linewidth]{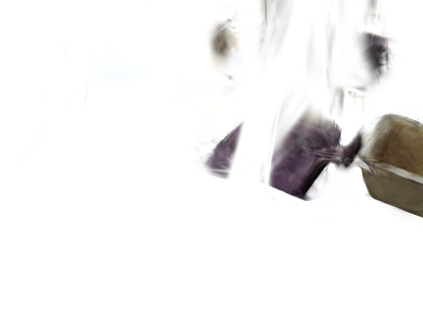} & \includegraphics[width=.134\linewidth]{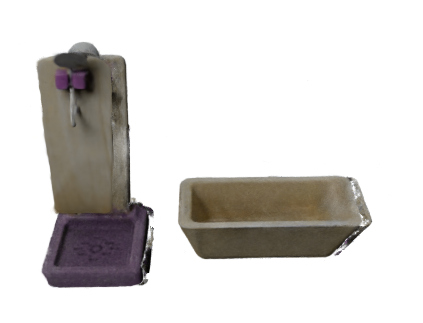} & \includegraphics[width=.134\linewidth]{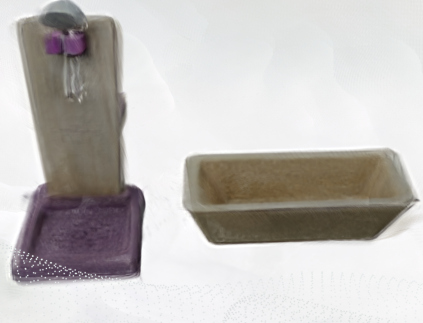} & \includegraphics[width=.134\linewidth]{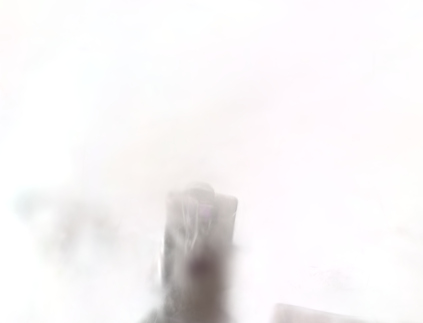} & \includegraphics[width=.134\linewidth]{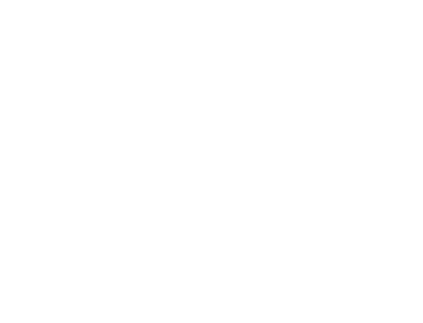} \\[3pt]
\end{tabular}
\endgroup

%% file: src/4_revision_results.tex
\subsection{Which components account for the improvement?}
\label{sec:synthetic_benchmark}
Comparisons with external methods combine differences in initialization, camera models, and optimization. To isolate the effect of angle refinement, we compare variants using the same input images, VGGT predictions, trainer, and training budget. The key comparison holds the estimated angles fixed in one variant and optimizes them jointly with the scene in the other.

\paragraph{The shared rig alone is not enough.}
With estimated angles held fixed, the shared rig achieves 11.99dB mean FG-PSNR, below the 12.42dB of independent cameras (\tabref{tab:synthetic_30k}). Their relative rotation errors are also similar: $12.46^\circ$ and $12.11^\circ$, respectively. This shows that the shared motion constraint alone does not compensate for inaccurate angles. Under regular capture, uniform angles achieve 26.29dB, compared with 12.04dB for frozen image-derived angles in the same rig (\tabref{tab:synthetic_robustness}). Accurate angular spacing is therefore important even when all views share one motion model.

\input{tables/tex/synthetic_robustness}

\input{figures/tex/synthetic_qual}

\paragraph{Joint angle refinement improves reconstruction.}
Jointly refining the estimated angles raises mean FG-PSNR to 19.88dB, an average improvement of 7.88dB over keeping them fixed. Refinement improves the mean in every capture condition (\tabref{tab:synthetic_robustness}), and the median improvement is 7.00dB. Mean object-crop LPIPS decreases from 0.281 to 0.142, while silhouette IoU increases from 0.718 to 0.860 (\tabref{tab:synthetic_30k}). Relative rotation error also decreases, and angular-spacing RMSE falls from $10.10^\circ$ to $8.53^\circ$. The qualitative examples show clearer silhouettes and more complete object parts after refinement (\figref{fig:synthetic_qual}).

\paragraph{Initialization and the shared rig both matter.}
We use two additional ablations to test the roles of image-derived initial angles and the shared rig (\tabref{tab:controls}). When initialized with uniform angles, refinement improves FG-PSNR by 1.96dB over keeping those angles fixed, but remains 5.80dB below OC-GS. This result supports the use of image-derived angles as the starting point for refinement.

For independent cameras, increasing both camera learning rates tenfold improves FG-PSNR by only 0.23dB, leaving a 6.92dB gap to OC-GS. At the tested settings, faster camera optimization does not recover the benefit of refining image-derived angles within the shared rig. For these ablations, we first average over valid conditions for each object, then average equally across objects.

\input{tables/tex/controls}

\paragraph{The gain persists across evaluation and loss choices.}
The average refinement gain remains between 6.9 and 8.1dB when we change how alignment failures are handled, apply stricter camera-alignment criteria, or exclude target views close to the training views (\appref{app:alignment} and \appref{app:followup}). In a separate experiment with three $L_1$/SSIM loss weightings, refinement improves mean FG-PSNR in all four capture conditions under every weighting. Average gains range from 4.88 to 7.68dB (\appref{app:followup}).

\paragraph{Refinement narrows the gap to true angles.}
Refinement reduces, but does not eliminate, the gap to the true-angle reference. With twelve irregular views, uniform angles produce FG-PSNR 18.68dB below the reference. Image-derived angles close 5.65dB of this gap, and joint refinement adds another 9.08dB. With six views, refinement still provides a median improvement of 4.23dB over frozen estimates, although mean quality remains below the true-angle reference.

%% file: tables/tex/synthetic_robustness.tex
\begin{table}[t]
\centering\footnotesize\setlength{\tabcolsep}{4pt}
\caption{\textbf{Effect of the shared rig and angle refinement.} FG-PSNR with the same trainer and VGGT predictions. Uniform angles assume equal spacing. Frozen angles retain image-derived estimates, while OC-GS refines them. The reference receives true angles. Means use the same valid objects per condition across all five variants and three seeds. Bold and underline indicate the best and second-best pose-free variants.}
\label{tab:synthetic_robustness}
\input{tables/revision_ablation_conditions}
\end{table}

%% file: tables/revision_ablation_conditions.tex
\begin{tabular}{lrrrr}
\toprule
& \multicolumn{4}{c}{FG-PSNR (dB) $\uparrow$} \\
\cmidrule(lr){2-5}
Configuration & Regular-12 & Irregular-12 & Irregular-8 & Irregular-6 \\
\midrule
No shared rig & 13.07 & \second{12.48} & \second{12.47} & \second{11.61} \\
Rig + uniform angles & \best{26.29} & 6.54 & 6.65 & 6.28 \\
Rig + frozen estimated angles & 12.04 & 12.19 & 12.13 & 11.61 \\
\ourrow OC-GS & \second{22.75} & \best{21.26} & \best{19.36} & \best{15.83} \\
\midrule
Rig + true angles (reference) & 26.16 & 25.22 & 22.24 & 18.19 \\
\midrule
\multicolumn{5}{@{}l}{\emph{Gain from refining the estimated angles}} \\
Mean (dB) & +10.71 & +9.08 & +7.23 & +4.22 \\
\bottomrule
\end{tabular}

%% file: figures/tex/synthetic_qual.tex
\begin{figure}[hbt]
\centering
\includegraphics[width=\linewidth]{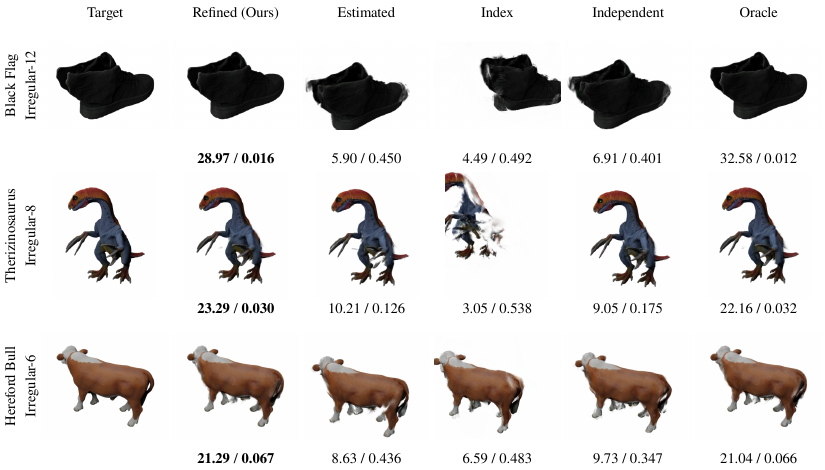}
\caption{\textbf{Selected examples of improvement with OC-GS.} Black Flag (12 views), Therizinosaurus (8), and Hereford Bull (6) were selected to illustrate reconstruction gains. Each row uses the same target view and crop across methods. Scores report FG-PSNR (dB) / object-crop LPIPS. Bold indicates the best valid pose-free result.}
\label{fig:synthetic_qual}
\end{figure}

%% file: tables/tex/controls.tex
\begin{table}[hbt]
\centering\footnotesize\setlength{\tabcolsep}{2.5pt}
\caption{\textbf{Refinement without image-derived angles or without the rig.} Uniform angles $+$ refinement removes image-derived angular initialization. Bold marks the best mean per column using FG-PSNR (dB).}
\label{tab:controls}
\input{tables/controls}
\end{table}

%% file: tables/controls.tex
\begin{tabular}{@{}lrrrr@{}}
\toprule
Configuration & Regular-12 $\uparrow$ & Irregular-12 $\uparrow$ & Irregular-8 $\uparrow$ & Irregular-6 $\uparrow$ \\
\midrule
Uniform angles $+$ refinement & \best{23.39} & 12.78 & 10.65 & 7.57 \\
Independent cameras (higher rate) & 13.30 & 12.75 & 12.70 & 11.83 \\
OC-GS & 22.75 & \best{21.26} & \best{19.36} & \best{15.83} \\
\bottomrule
\end{tabular}

%% file: tables/tex/ocgs_initializers.tex
\begin{table}[hbt]
\centering\footnotesize\setlength{\tabcolsep}{5pt}
\caption{\textbf{OC-GS with different initializers.} We show the performance of different initializers and report them with the mean FG-PSNR (dB $\uparrow$). The best is bolded, which is from VGGT.}
\label{tab:ocgs_initializers}
\input{tables/ocgs_initializers}
\end{table}

%% file: tables/ocgs_initializers.tex
\begin{tabular}{@{}lrrrr@{}}
\toprule
OC-GS initializer & Regular-12 & Irregular-12 & Irregular-8 & Irregular-6 \\
\midrule
VGGT & \textbf{22.75} & \textbf{21.26} & \textbf{19.36} & \textbf{15.83} \\
AnySplat & 19.88 & 20.28 & 18.30 & 15.10 \\
$\pi^3$ & 19.64 & 20.45 & 18.61 & 15.19 \\
VGGT-$\Omega$ & 21.19 & 20.13 & 18.87 & 15.57 \\
Depth Anything 3 & 20.24 & 19.77 & 18.19 & 15.28 \\
CUT3R & 17.26 & 18.45 & 15.71 & 14.70 \\
MapAnything & 7.81 & 8.39 & 6.70 & 5.82 \\
DUSt3R & 13.35 & 15.42 & 14.36 & 11.10 \\
MASt3R-SfM & 17.89 & 14.43 & 8.46 & 7.07 \\
\midrule
Index angles (VGGT geometry/rig) & 15.55 & 11.33 & 10.13 & 9.28 \\
\bottomrule
\end{tabular}

%% file: tables/tex/init_sources.tex
\begin{table}[hbt]
\centering\footnotesize\setlength{\tabcolsep}{5pt}
\caption{\textbf{Pose initialization and camera models on real captures.} Mean FG-PSNR (dB), evaluated using each reconstruction's cameras. The named methods provide initial poses. The OC-GS rig keeps its azimuths fixed, while independent cameras are optimized from the predictions. Bold and underline indicate the best and second-best predictors within each block at the displayed precision.}
\label{tab:init_sources}
\input{tables/init_sources}
\end{table}

%% file: tables/init_sources.tex
\begin{tabular}{@{}lrrrr@{}}
\toprule
Configuration / training views & 24 & 17 & 13 & 7 \\
\midrule
\multicolumn{5}{@{}l}{\emph{OC-GS shared rig, initialized azimuths frozen}} \\
\quad AnySplat azimuths & \best{22.62} & \best{21.53} & \best{19.50} & \best{15.73} \\
\quad $\pi^3$ azimuths & \second{21.82} & \second{20.91} & \best{19.50} & \second{15.72} \\
\quad Index angles (no network) & 21.20 & 17.12 & 16.40 & 13.56 \\
\midrule
\multicolumn{5}{@{}l}{\emph{Per-view cameras, optimized from}} \\
\quad $\pi^3$ & \second{20.45} & \second{18.70} & \best{16.75} & \best{14.03} \\
\quad AnySplat & \best{20.67} & \best{18.91} & \second{15.36} & \second{13.82} \\
\quad Depth Anything 3 & 18.52 & 15.49 & 14.30 & 13.53 \\
\quad VGGT & 17.52 & 16.08 & 13.97 & 11.56 \\
\quad VGGT-$\Omega$ & 17.51 & 15.89 & 14.87 & 10.38 \\
\quad CUT3R & 16.73 & 14.93 & 12.66 & 9.33 \\
\quad MASt3R-SfM & 17.23 & 13.96 & 10.71 & 9.85 \\
\quad MapAnything & 16.38 & 13.40 & 12.17 & 9.15 \\
\quad DUSt3R & 14.99 & 13.45 & 11.05 & 7.33 \\
\quad Index angles (no network) & 20.79 & 17.08 & 15.83 & 12.12 \\
\bottomrule
\end{tabular}

%% file: tables/tex/followup_dose.tex
\begin{table}[hbt]
\centering\footnotesize\setlength{\tabcolsep}{4pt}
\caption{\textbf{Angle-error experiment.} RMS spacing error is the root-mean-square angular noise in degrees after removing a common rotation. \texttt{Frozen} keeps the supplied angles fixed, while \texttt{Refined} optimizes them. $\Delta$ reports the refinement gain.}
\label{tab:followup_dose}
{\renewcommand{\arraystretch}{0.92}\input{tables/followup_dose}}
\end{table}

%% file: tables/followup_dose.tex
\begin{tabular}{@{}lrrr@{}}
\toprule
RMS spacing error & Frozen $\uparrow$ & Refined $\uparrow$ & $\Delta$ (dB) \\
\midrule
$0^\circ$ (exact angles) & 26.04 & 22.88 & $-3.16$ \\
$0.5^\circ$ & 18.44 & 23.19 & $+4.75$ \\
$1^\circ$ & 15.89 & 22.92 & $+7.03$ \\
$2^\circ$ & 13.50 & 22.60 & $+9.10$ \\
$3^\circ$ & 12.36 & 22.71 & $+10.35$ \\
$10^\circ$ & 9.08 & 20.33 & $+11.25$ \\
\bottomrule
\end{tabular}

%% file: tables/tex/cost.tex
\begin{table}[htbp]
\centering\scriptsize\setlength{\tabcolsep}{4pt}
\caption{\textbf{Reconstruction quality, runtime, and peak GPU memory.} Quality uses valid benchmark runs. Cost uses separate profiling runs. VGGT inference is excluded from runtime and memory for OC-GS and Nerfacto. Bold and underline indicate the best and second-best values at the displayed precision.}
\label{tab:cost}
{\renewcommand{\arraystretch}{0.92}\input{tables/cost_matched_main}}
\vspace{-4pt}
\end{table}

%% file: tables/cost_matched_main.tex
\begin{tabular}{@{}llrr@{}}
\toprule
Method & Type & Time (min) $\downarrow$ & Peak VRAM (GB) $\downarrow$ \\
\midrule
\ourrow OC-GS & Per-scene  & 16.2 & \textbf{1.51} \\
RotGS (index schedule) & Per-scene & 20.8 & \underline{2.20} \\
Nerfacto + VGGT poses & Per-scene & 23.7 & 3.05 \\
InstantSplat & Per-scene & \underline{2.5} & 6.69 \\
AnySplat & Feed-forward & \textbf{0.5} & 8.31 \\
SPFSplat & Feed-forward & \textbf{0.5} & 6.06 \\
\bottomrule
\end{tabular}

%% file: src/5_limitations.tex
\section{Limitations}
\label{sec:limitations}
OC-GS assumes a fixed axis with minimal axis noise (\appref{app:rotgs_axis_audit}) and depends on initialization(VGGT). However, we expect the future models to improve reconstruction.

%% file: src/6_conclusion.tex
\section{Conclusion}
\label{sec:conclusion}
OC-GS reconstructs objects from sparse, irregular turntable images by jointly optimizing image-derived angles and Gaussian geometry under a shared motion model. It outperforms the evaluated pose-free GS baselines in all three irregular capture conditions. With the same trainer and initialization, angle refinement adds 9.08dB FG-PSNR with twelve irregular views and 4.22dB with just six views, compared with keeping the estimated angles fixed (\tabref{tab:synthetic_robustness}). Moreover, OC-GS uses 31\% less peak GPU memory and takes 22\% less fitting time than RotGS, excluding VGGT inference (\tabref{tab:cost}). Future work will adapt refinement to initialization quality and evaluate its use in plant phenotyping as a downstream task.

%% file: src/2_related.tex
\section{Related work}
\label{sec:related}
\paragraph{Gaussian reconstruction from sparse images.}
3DGS~\citep{kerbl2023gaussian} and 2DGS~\citep{huang20242d} optimize Gaussian scenes from images with known poses, often estimated by COLMAP~\citep{schoenberger2016sfm}. InstantSplat~\citep{fan2024instantsplat} combines learned geometry with joint Gaussian and camera optimization. FreeSplatter~\citep{xu2024freesplatter}, AnySplat~\citep{jiang2025anysplat}, and SPFSplat~\citep{huang2025spfsplat} reconstruct Gaussian scenes without supplied poses. OC-GS uses the turntable motion to constrain reconstruction: all views share one camera, axis, and pivot, with a separate angle for each image.

\paragraph{Learning geometry and camera poses.}
DUSt3R~\citep{wang2024dust3r} predicts dense point maps, and MASt3R~\citep{leroy2024mast3r} adds matching features for geometric alignment. VGGT~\citep{wang2025vggt} predicts cameras and scene geometry jointly. $\pi^3$~\citep{wang2025pi3} removes the need for a designated reference view, while CUT3R~\citep{wang2025cut3r} maintains a persistent state across images. Depth Anything 3~\citep{lin2025depthanything3}, MapAnything~\citep{keetha2025mapanything}, and VGGT-$\Omega$~\citep{wang2026vggtomega} provide additional sources of geometry and camera estimates. OC-GS converts predicted cameras into initial turntable angles and refines them using the captured images.

\paragraph{Joint camera and scene refinement.}
NeRF--~\citep{wang2021nerfmm}, BARF~\citep{lin2021barf}, and SCNeRF~\citep{jeong2021scnerf} jointly optimize camera poses and scene radiance. CamP~\citep{park2023camp} improves camera optimization through preconditioning. SPARF~\citep{truong2023sparf} uses correspondence and depth consistency, while PoRF~\citep{bian2024porf} learns a pose residual field. NeRS~\citep{zhang2021ners}, SAMURAI~\citep{boss2022samurai}, and NeROIC~\citep{kuang2022neroic} reconstruct objects from image collections with uncertain cameras. OC-GS constrains all camera updates to a shared rotation trajectory throughout joint scene and motion optimization.

\paragraph{Turntable geometry.}
Earlier work recovered shape and camera motion using silhouettes, circular-motion constraints, and conic fitting~\citep{fitzgibbon1998turntable,mendonca2001epipolar,wong2001silhouettes,jiang2003conic,hernandez2007silhouette}. RotGS~\citep{kim2026rotgs} combines a shared rotation axis with optical-flow alignment for Gaussian reconstruction. Its uniform index schedule and interpolated endpoint correction cannot represent arbitrary per-frame gaps. OC-GS initializes nonuniform angles from images and refines each angle around a shared axis.

%% file: src/7_appendix.tex
\noindent
This appendix provides the experimental details and supporting results. Section~\ref{app:gs_protocol} describes data and training, and Section~\ref{app:alignment} defines camera alignment and evaluation metrics. Section~\ref{app:corrections} reports complete baseline comparisons and computational costs. Sections~\ref{app:real_profile} and~\ref{app:followup} cover real captures and sensitivity tests. Section~\ref{app:initablation} examines initialization and sources of error in the baselines.

\section{Benchmark and implementation details}
\label{app:gs_protocol}\label{app:protocol}\label{app:statistics}

\paragraph{Data and schedules.}
The synthetic benchmark contains 100 Google Scanned Objects rendered at $512\times512$ pixels. Real captures come from the dataset released with RotGS~\citep{kim2026rotgs}. Their settings are given in \appref{app:real_profile}. Each synthetic capture condition has 36 calibrated target views at $35^\circ$ elevation, spaced every $10^\circ$ from an initial azimuth of $0.371^\circ$.

Regular training views are equally spaced around a full turn. For irregular captures, we sample angular gaps from a lognormal distribution with log-standard deviation 0.8. We smooth neighboring gaps with weights 0.25, 0.5, and 0.25, then rescale them to sum to $360^\circ$. Sparse captures select evenly spaced indices from the 24-image sequence, rounding down. These subsets are not nested. We exclude targets within $0.05^\circ$ of a training view. Methods evaluated without known poses use reference extrinsics only for evaluation alignment. Section~\ref{app:corrections} specifies which methods receive or estimate intrinsics.

\paragraph{Optimization.}
The primary component experiments use the same VGGT predictions and 30k training steps. They use the released trainer's default parameters. The photometric loss is pure $L_1$. We disable the silhouette, depth, consistency, normal, and distortion losses. Angle refinement uses a learning rate of 0.001, subtracts the mean correction, and applies no angle penalty. In the independent-camera variant with a higher learning rate, both camera rates increase tenfold, to 0.001 for translation and 0.0001 for rotation.

\paragraph{Angle initialization.}
We first fit a plane to the VGGT camera centers by least squares, then fit a circle in that plane. Consecutive angle changes determine the rotation direction after accounting for wraparound, and the first image defines zero rotation. The fit uses every predicted center without outlier rejection. Refinement does not enforce the initial angle ordering.

\paragraph{Random seeds.}
The primary component experiments and optimized external comparisons use three random seeds. Supporting studies use one seed unless stated otherwise.

\FloatBarrier
\section{Camera alignment and evaluation}
\label{app:alignment}
\paragraph{Aligning cameras using training views.}
We align reference cameras to the reconstructed scene with one rotation, scale, and translation. We first fit the rotation using reference and learned training-camera orientations, then fit scale and translation to their centers by least squares. All target cameras receive the same transform. 

\paragraph{Alignment checks.}
We cross-check the orientation-based alignment using Umeyama's camera-center fit~\citep{umeyama1991least}. We flag center errors above 5\% of the orbit radius, rotation disagreements above $2^\circ$, or scale disagreements above 2\%. These flags do not change the main scores. Restricting evaluation to orientation-based alignments that are stable across all variants and seeds gives a refinement gain of 8.05dB.

\paragraph{Metrics.}
FG-PSNR measures RGB error on target foreground pixels. Object-crop LPIPS uses AlexNet v0.1 with RGB values scaled to $[-1,1]$. Its crop is the target-mask bounding box, padded by 15\% on each side and clipped to the image boundary. Silhouette IoU compares rendered opacity with target masks after thresholding both at 0.5. RRE averages the angular discrepancy between learned and reference relative rotations over all training-camera pairs.

\input{tables/tex/synthetic_30k}

\FloatBarrier
\section{External baselines and complete comparisons}
\label{app:corrections}\label{app:metric_provenance}\label{app:full_comparison}

Feed-forward methods make one prediction per case. AnySplat and InstantSplat estimate intrinsics. OC-GS, SPFSplat, and Nerfacto with VGGT cameras receive known intrinsics. The calibrated Nerfacto and Splatfacto references receive true intrinsics and extrinsics. InstantSplat uses MASt3R initialization~\citep{leroy2024mast3r}, while calibrated Splatfacto starts from random Gaussians. SPFSplat uses weights trained on RE10K~\citep{zhou2018stereo} and expects ten context views. Additional experiments are described in \appref{app:baseline_failure}.

\input{tables/tex/external_baselines}

\input{tables/tex/external_paired}

\input{tables/tex/external_pooled}

\paragraph{Cost and training budget.}
We profile the first eight objects in the dataset list under regular-12 and irregular-6 capture on an otherwise idle RTX A5000. Runtime is measured from launch to exit. GPU memory is polled every 0.5s. Peak usage includes the CUDA context and subtracts idle usage. AnySplat, SPFSplat, and InstantSplat measurements include geometry inference. OC-GS and Nerfacto use precomputed VGGT predictions. Quality scores use valid archived cases, while cost averages all 16 cases per method. Table~\ref{tab:cost_full} reports costs for the references with true cameras.

OC-GS quality improves between 6k and 30k training steps (\tabref{tab:cost_convergence}). On the same objects with valid alignment, the benchmark and separate profiling runs achieve mean FG-PSNR of 15.74 and 15.42dB at 30k steps. Their mean absolute paired difference is 0.51dB.

\input{tables/tex/cost_full}

\input{tables/tex/cost_convergence}

\FloatBarrier
\section{Real-capture experiments}
\label{app:real_profile}\label{sec:rotgs_control}\label{sec:efficiency}

We evaluate nine captures with 24, 17, 13, and 7 training views, using 30k steps and a pure $L_1$ loss. The fixed-angle and refined-angle variants use the earlier implementation with identical AnySplat angles, input images, target views, and camera conventions. Angle refinement is the only difference. Target cameras are estimated rather than independently calibrated.

\input{tables/tex/real_refined}

\FloatBarrier
\section{Sensitivity and component experiments}
\label{app:followup}
These experiments examine when refinement helps and how its gains depend on the training loss and target-view selection.

\paragraph{Sensitivity to angle errors.}
We perturb the true angles while keeping regular-12 input images and VGGT predictions fixed. For each nonzero error level, we average two noise samples per object. The finer error levels in \tabref{tab:followup_crossover} place the transition from harmful to helpful refinement between $0.1^\circ$ and $0.2^\circ$ under these settings. The supplied angles also affect initialization, so comparisons across error levels include this effect.

\input{tables/tex/followup_crossover}

With exact initial angles, both variants achieve similar training photometric errors, but refinement reduces target-view quality.

\paragraph{Photometric loss.}
Refinement improves reconstruction under different $L_1$/SSIM weightings. A separate three-seed study gives mean FG-PSNR gains of 7.68dB with pure $L_1$, 5.99dB with weights $0.8/0.2$, and 4.88dB with weights $0.6/0.4$. Every capture condition improves. Tests with weights $0.9/0.1$, $0.4/0.6$, and pure SSIM also show positive gains.

\paragraph{Target-view separation.}
Excluding target views within $5^\circ$ or $10^\circ$ of a training angle gives refinement gains of 7.16 and 6.91dB, respectively. The benefit therefore persists when evaluation excludes views close to the inputs.

\FloatBarrier
\section{Initialization and baseline behavior}
\label{app:initablation}
We validate how camera estimates and field settings affect the component and baseline results.

\subsection{RotGS: camera fit and axis sensitivity}
\label{app:rotgs_axis_audit}
For regular-12 capture, the trained RotGS scene achieves 33.35dB FG-PSNR on training views using its estimated cameras and 14.24dB using aligned reference cameras. We exclude the repeated endpoint image.

The axis experiment keeps the initial geometry and the 30k-step training budget fixed while allowing the pivot to be optimized. Fixing the axis to its true value increases mean FG-PSNR from 15.97 to 29.45dB, with every tested object improving. Axis estimation can therefore limit reconstruction even under regular capture.

\input{src/7_baseline_failure}

\input{src/7_nerfacto_failure}

%% file: tables/tex/synthetic_30k.tex
\begin{center}
\begin{minipage}{\linewidth}
\centering\footnotesize\setlength{\tabcolsep}{3pt}
\captionof{table}{\textbf{Pooled component comparisons.} FG-PSNR, object LPIPS, and mask IoU use the same valid objects across variants. RRE includes all reconstructions, including those with invalid alignment. Bold and underline indicate the best and second-best pose-free variants. The true-angle reference is unranked.}
\label{tab:synthetic_30k}
\input{tables/revision_ablation_main}
\end{minipage}
\end{center}

%% file: tables/revision_ablation_main.tex
\begin{tabular}{lrrrr}
\toprule
Configuration & FG-PSNR (dB) $\uparrow$ & Obj. LPIPS $\downarrow$ & IoU $\uparrow$ & RRE ($^\circ$) $\downarrow$ \\
\midrule
Independent cameras & \second{12.42} & \second{0.263} & \second{0.749} & \second{12.11} \\
Rig: index & 11.70 & 0.427 & 0.474 & 22.15 \\
Rig: estimated & 11.99 & 0.281 & 0.718 & 12.46 \\
OC-GS & \best{19.88} & \best{0.142} & \best{0.860} & \best{10.25} \\
\midrule
Rig: oracle & 23.04 & 0.106 & 0.886 & 6.33 \\
\bottomrule
\end{tabular}

%% file: tables/tex/external_baselines.tex
\begin{center}
\begin{minipage}{\linewidth}
\centering\scriptsize\setlength{\tabcolsep}{3pt}
\captionof{table}{\textbf{Complete external comparison.} Results on the full rendered dataset, averaged over seeds and then over each method's valid cases. Methods in the lower block receive true cameras. Best scores are not highlighted because the sets of valid cases differ.}
\label{tab:external_baselines}
\input{tables/external_baselines_matched}
\end{minipage}
\end{center}

%% file: tables/external_baselines_matched.tex
\begin{tabular}{@{}lrrrr@{}}
\toprule
Method & Regular-12 $\uparrow$ & Irregular-12 $\uparrow$ & Irregular-8 $\uparrow$ & Irregular-6 $\uparrow$ \\
\midrule
RotGS (index schedule) & 13.96 & 5.64 & 5.17 & 5.17 \\
Nerfacto + VGGT poses & 10.14 & 11.55 & 8.14 & 6.30 \\
AnySplat & 11.61 & 12.41 & 12.05 & 11.46 \\
InstantSplat & 5.06 & 4.51 & 4.98 & 4.72 \\
SPFSplat & 3.73 & 3.81 & 3.83 & 3.85 \\
\ourrow OC-GS (ours) & 22.75 & 21.26 & 18.18 & 15.39 \\
\midrule
Nerfacto (calibrated) & 13.02 & 15.64 & 10.96 & 6.61 \\
Splatfacto (calibrated) & 24.64 & 23.31 & 14.60 & 9.28 \\
\bottomrule
\end{tabular}

%% file: tables/tex/external_paired.tex
\begin{center}
\begin{minipage}{\linewidth}
\centering\scriptsize\setlength{\tabcolsep}{3pt}
\captionof{table}{\textbf{Paired FG-PSNR differences (dB).} OC-GS minus baseline on shared valid cases. Means use only cases valid for both methods.}
\label{tab:external_paired}
\input{tables/external_baselines_paired}
\end{minipage}
\end{center}

%% file: tables/external_baselines_paired.tex
\begin{tabular}{@{}lrrrr@{}}
\toprule
Comparison & Regular-12 & Irregular-12 & Irregular-8 & Irregular-6 \\
\midrule
OC-GS $-$ RotGS (index schedule) & +8.80 & +15.57 & +13.01 & +10.25 \\
OC-GS $-$ Nerfacto + VGGT poses & +12.28 & +9.40 & +10.57 & +9.53 \\
OC-GS $-$ AnySplat & +11.14 & +9.53 & +6.73 & +4.11 \\
OC-GS $-$ InstantSplat & +17.69 & +17.43 & +14.50 & +11.60 \\
OC-GS $-$ SPFSplat & +19.02 & +17.32 & +14.26 & +10.71 \\
\bottomrule
\end{tabular}

%% file: tables/tex/external_pooled.tex
\begin{center}
\begin{minipage}{\linewidth}
\centering\scriptsize\setlength{\tabcolsep}{3pt}
\captionof{table}{\textbf{Pooled paired comparisons.} Positive FG-PSNR and negative object LPIPS favor OC-GS. Lower block receives true cameras.}
\label{tab:external_pooled}
\input{tables/submission_baselines_pooled}
\end{minipage}
\end{center}

%% file: tables/submission_baselines_pooled.tex
\begin{tabular}{@{}lrrr@{}}
\toprule
Baseline & Paired cases & $\Delta$ FG-PSNR & $\Delta$ Obj. LPIPS \\
\midrule
RotGS (index + endpoint) & 98 & +11.89 & -0.358 \\
Nerfacto + VGGT & 90 & +10.45 & -0.325 \\
AnySplat & 95 & +7.94 & -0.440 \\
InstantSplat & 92 & +15.41 & -0.481 \\
SPFSplat & 70 & +15.76 & -0.482 \\
\midrule
Nerfacto (true cameras) & 98 & +7.86 & -0.221 \\
Splatfacto (true cameras) & 98 & +1.43 & -0.103 \\
\bottomrule
\end{tabular}

%% file: tables/tex/cost_full.tex
\begin{center}
\begin{minipage}{\linewidth}
\centering\scriptsize\setlength{\tabcolsep}{4pt}
\captionof{table}{\textbf{Calibrated references for the cost benchmark.}
Both methods use true cameras. Quality uses archived runs and
cost uses separate GPU profiling runs.}
\label{tab:cost_full}
\input{tables/cost_calibrated}
\end{minipage}
\end{center}

%% file: tables/cost_calibrated.tex
\begin{tabular}{@{}llrrr@{}}
\toprule
Method & Type & FG-PSNR (dB) $\uparrow$ & Time (min) $\downarrow$ & Peak VRAM (GiB) $\downarrow$ \\
\midrule
Nerfacto (calibrated) & Per-scene & 11.40 & 13.7 & 2.27 \\
Splatfacto (calibrated) & Per-scene & 16.20 & 8.0 & 1.83 \\
\bottomrule
\end{tabular}

%% file: tables/tex/cost_convergence.tex
\begin{center}
\begin{minipage}{\linewidth}
\centering\scriptsize\setlength{\tabcolsep}{4pt}
\captionof{table}{\textbf{Quality versus training budget.} OC-GS on eight objects
under regular-12 and irregular-6 capture, using an RTX A5000.
Time and peak VRAM average all profiled cases. FG-PSNR uses the same objects with valid alignment at every budget.}
\label{tab:cost_convergence}
\input{tables/cost_convergence}
\end{minipage}
\end{center}

%% file: tables/cost_convergence.tex
\begin{tabular}{@{}lrrr@{}}
\toprule
OC-GS budget & Time (min) & FG-PSNR (dB) $\uparrow$ & Peak VRAM (GiB) \\
\midrule
6k & 3.3 & 13.07 & 1.47 \\
12k & 6.4 & 13.72 & 1.48 \\
30k & 16.2 & 15.42 & 1.51 \\
\bottomrule
\end{tabular}

%% file: tables/tex/real_refined.tex
\begin{center}
\begin{minipage}{\linewidth}
\centering\footnotesize\setlength{\tabcolsep}{3pt}
\captionof{table}{\textbf{Real freeze-versus-refine results.} Means across nine captures. FG-PSNR and paired gain are in dB. LPIPS lists frozen then refined. Pooled quantities are paired differences, not averages of absolute quality across view counts.}
\label{tab:real_refined}
\input{tables/real_refined}
\end{minipage}
\end{center}

%% file: tables/real_refined.tex
\begin{tabular}{@{}lrrrrr@{}}
\toprule
Training views & Frozen $\uparrow$ & Refined $\uparrow$ & $\Delta$ (dB) & Median & Obj.\ LPIPS $\downarrow$ \\
\midrule
24 & 24.30 & \best{25.04} & $+0.74$ & $+0.62$ & $0.102\!\rightarrow\!0.093$ \\
17 & 22.84 & \best{23.87} & $+1.03$ & $+0.84$ & $0.116\!\rightarrow\!0.099$ \\
13 & 20.71 & \best{21.18} & $+0.47$ & $+0.21$ & $0.144\!\rightarrow\!0.134$ \\
7 & 16.46 & \best{17.04} & $+0.57$ & $+0.25$ & $0.224\!\rightarrow\!0.206$ \\
\midrule
All 36 cases & \textendash{} & \textendash{} & $+0.70$ & $+0.60$ & \textendash{} \\
\bottomrule
\end{tabular}

%% file: tables/tex/followup_crossover.tex
\begin{center}
\begin{minipage}{\linewidth}
\centering\footnotesize\setlength{\tabcolsep}{3pt}
\captionof{table}{\textbf{Fine angle-error study.} Regular-12 images with perturbed true angles. RMS spacing error measures per-view angle error after removing a common rotation. FG-PSNR and the paired refinement gain are in dB. Bold marks the better mean.}
\label{tab:followup_crossover}
\input{tables/followup_crossover}
\end{minipage}
\end{center}

%% file: tables/followup_crossover.tex
\begin{tabular}{@{}lrrr@{}}
\toprule
RMS spacing error & Frozen $\uparrow$ & Refined $\uparrow$ & $\Delta$ (dB) \\
\midrule
$0^\circ$ (exact angles) & \best{26.04} & 22.88 & $-3.16$ \\
$0.1^\circ$ & \best{24.22} & 23.22 & $-1.00$ \\
$0.2^\circ$ & 21.93 & \best{23.19} & $+1.25$ \\
$0.3^\circ$ & 20.46 & \best{23.03} & $+2.58$ \\
$0.5^\circ$ & 18.44 & \best{23.19} & $+4.75$ \\
\bottomrule
\end{tabular}

%% file: src/7_baseline_failure.tex
\subsection{Image-based baselines: camera consistency}
\label{app:baseline_failure}
We examine AnySplat, InstantSplat, and SPFSplat reconstructions under regular-12 and irregular-6 capture. We hold each scene fixed and replace its inferred intrinsics, then extrinsics, with calibrated values. Table~\ref{tab:baseline_failure} reports the resulting changes in training-image quality. These changes measure camera--scene consistency, including the effects of scene alignment.

\input{tables/tex/baseline_failure}

Replacing the cameras reduces training-image quality for AnySplat and InstantSplat. SPFSplat already fits training images poorly before camera replacement. Using camera centers rather than orientations for alignment changes target FG-PSNR by $-0.03$ to $-0.21$dB and does not improve the results.

InstantSplat crops inputs from $512\times512$ to $512\times384$ but scales the exported vertical focal length by $4/3$. We correct this export on the cost-study subset under both capture conditions, keeping input images and the 1k-step budget fixed. Target FG-PSNR increases from 5.53 to 5.87dB. Neither variant receives a calibrated focal length, and the benchmark retains the released export. The correction explains only a small part of the gap to OC-GS.

%% file: tables/tex/baseline_failure.tex
\begin{center}
\begin{minipage}{\linewidth}
\centering\footnotesize\setlength{\tabcolsep}{4pt}
\captionof{table}{\textbf{Camera consistency with fixed reconstructions, FG-PSNR (dB).}
Results use a single random seed. Native train and the intrinsics
replacement drop $\Delta K$ use all objects per row. The extrinsics replacement drop $\Delta R,t$, target mean,
and center-fit target change average valid placements. Positive
drops indicate that replacing cameras lowers training-image quality.
A positive center change would favor the alternative placement.}
\label{tab:baseline_failure}
\input{tables/baseline_failure_diagnostics}
\end{minipage}
\end{center}

%% file: tables/baseline_failure_diagnostics.tex
\begin{tabular}{@{}llrrrrr@{}}
\toprule
Method & Capture & Native train & $\Delta K$ & $\Delta R,t$ & Target & $\Delta$ center \\
\midrule
AnySplat & Regular-12 & 16.05 & 1.52 & 2.91 & 11.61 & -0.08 \\
AnySplat & Irregular-6 & 17.94 & 3.05 & 3.01 & 11.46 & -0.17 \\
InstantSplat & Regular-12 & 14.11 & 4.70 & 4.22 & 5.06 & -0.16 \\
InstantSplat & Irregular-6 & 25.84 & 13.89 & 6.41 & 4.71 & -0.03 \\
SPFSplat & Regular-12 & 7.02 & 0.00 & 3.33 & 3.73 & -0.20 \\
SPFSplat & Irregular-6 & 8.23 & 0.00 & 4.43 & 3.85 & -0.21 \\
\bottomrule
\end{tabular}

%% file: src/7_nerfacto_failure.tex
Nerfacto can fit sparse training images without recovering the correct object geometry. Its default scene contraction allows density outside the object region, which can reproduce the training views but leads to limited visual-quality reconstructions from unseen viewpoints.